%% file: main.tex
\def\deanonymize{}
\documentclass[letterpaper, 10 pt, conference]{ieeeconf}

\IEEEoverridecommandlockouts   %
\usepackage[T1]{fontenc}
\usepackage{times}
\usepackage{amsmath,amssymb}
\usepackage{dsfont}
\usepackage{graphicx}
\usepackage{booktabs}
\usepackage{multirow}
\usepackage{xcolor}
\usepackage{cite}
\usepackage{url}
\usepackage{cuted}          %
\makeatletter
\newcommand{\captionof}[1]{\def\@captype{#1}\caption}
\makeatother

\newcommand{\method}{VideoReloc}                 %
\newcommand{\recall}{R@1\,m/10$^\circ$}         %

\title{\LARGE \bf
\method{}: Long-Term Indoor Video Relocalization\\
against a Kilobyte-Scale Semantic Scene Graph
}

\newif\ifanon
\anontrue
\ifdefined\deanonymize \anonfalse \fi
\ifanon\else
  \usepackage[hidelinks]{hyperref}
  \hypersetup{pdftitle={VideoReloc: Long-Term Indoor Video Relocalization against a Kilobyte-Scale Semantic Scene Graph},
              pdfauthor={Qianru Li, Xuyang Chen, Xuqin Wang, Zhenghao Zhang, Hongyi Luo, Tao Wu, Daniel Cremers, Lu Liu, Yanfeng Zhang}}
\fi

\ifanon
  \author{}
\else
  \author{Qianru Li$^{1,2}$, Xuyang Chen$^{1,2}$, Xuqin Wang$^{1,2}$,
  Zhenghao Zhang$^{1}$, Hongyi Luo$^{1,2}$\\
  Tao Wu$^{2}$, Daniel Cremers$^{1}$, Lu Liu$^{2}$ and Yanfeng Zhang$^{2}$%
  \thanks{$^{1}$Technical University of Munich, Munich, Germany.}%
  \thanks{$^{2}$Huawei Hilbert Research Center (Dresden), Germany.}%
  \thanks{Corresponding author: Yanfeng Zhang,\hfil\break {\tt\small zhang\_yanfeng\_3d@foxmail.com}.}%
  }
\fi

\begin{document}
\IEEEaftertitletext{\vspace{-36pt}}
\maketitle
\thispagestyle{empty}
\pagestyle{empty}

\input{sections/teaser_block}

\begin{abstract}
\input{sections/abstract}
\end{abstract}

\input{sections/introduction}
\input{sections/related_work}
\input{sections/method}
\input{sections/experiments}
\input{sections/conclusion}

\bibliographystyle{IEEEtran}
\bibliography{refs}

\end{document}

%% file: sections/teaser_block.tex
\begin{strip}
\centering
  \centering
  \begin{minipage}[b]{0.475\textwidth}
    \centering
    \includegraphics[width=\linewidth]{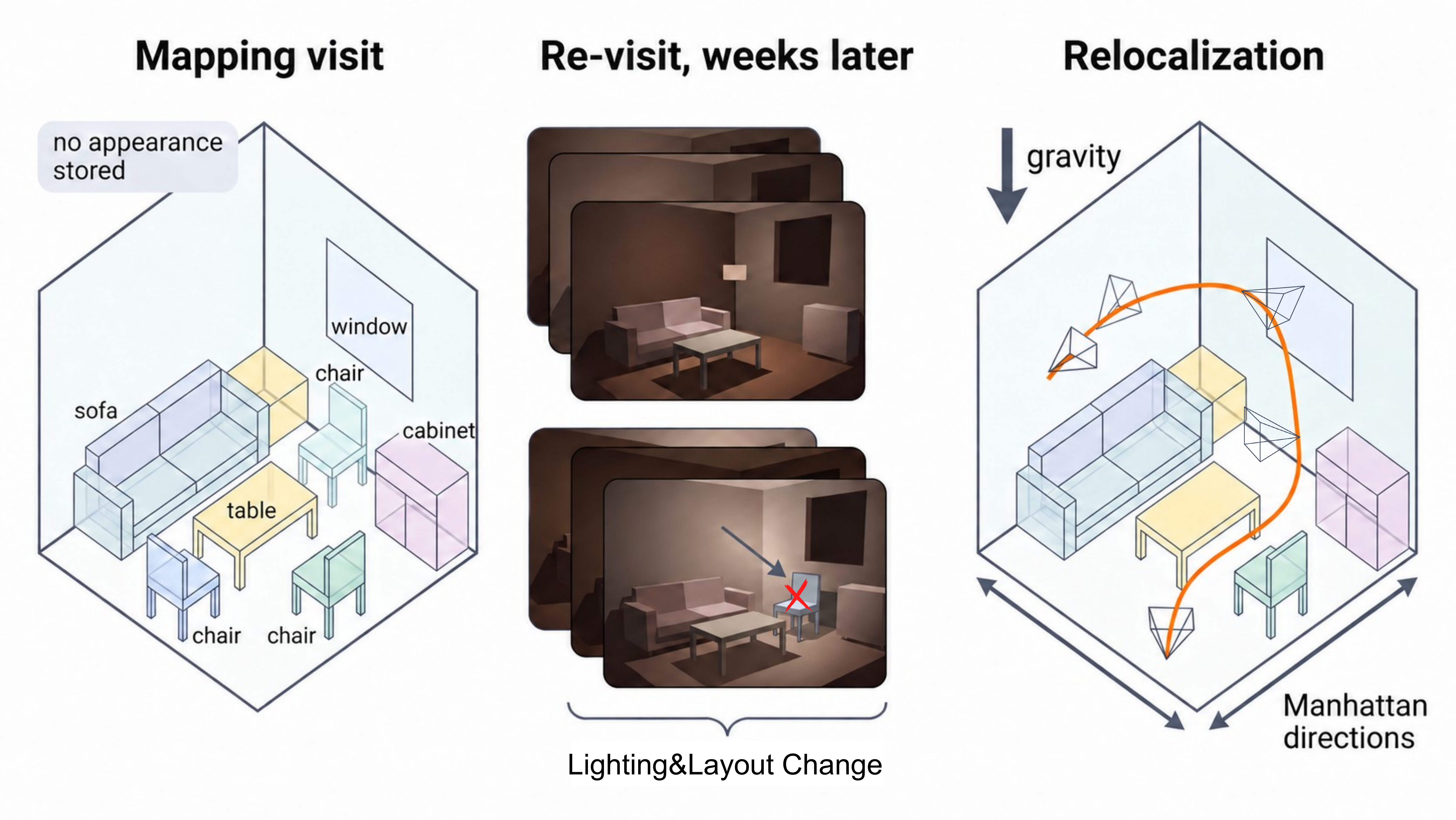}
  \end{minipage}\hfill
  \begin{minipage}[b]{0.505\textwidth}
    \centering
    \includegraphics[width=\linewidth]{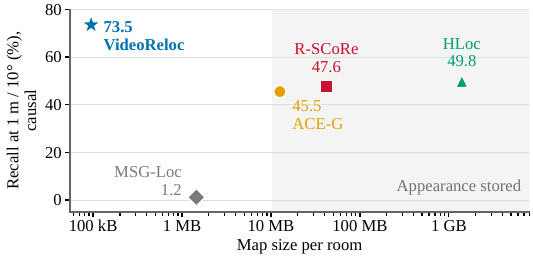}
  \end{minipage}
  \captionof{figure}{\textbf{Video relocalization with a compact semantic map.} \emph{Left:} the map stores class-labelled boxes for objects, walls and floors in about 100\,kB. A clip from a later RGB-D video is localized in the map's coordinate system after changes in lighting and furniture layout. \emph{Right:} map size versus recall at 1\,m/10$^\circ$ over all RIO10 re-scan frames. \method{} achieves the highest recall with the smallest map among these methods.}
  \label{fig:teaser}
\end{strip}

%% file: sections/abstract.tex
Given a compact semantic scene graph, long-term indoor video relocalization estimates a map-frame trajectory after lighting and furniture changes.
Visual methods rely on appearance and become unreliable under these changes; localizing one frame at a time from object classes and geometry instead leaves sparse, ambiguous evidence.
We introduce \method{}, whose adaptive clips use odometry to gather spatial evidence until object and motion criteria are met, adapting query length to the observed scene.
Its run-level decision rechecks conflicting placements using evidence accumulated across connected clips, stabilizing the trajectory beyond adjacent-clip tracking.
\emph{Hypothesis-first registration} proposes poses from object triplets and verifies each using clip-wide object centers and box surfaces.
\emph{Orientation-aware refinement} uses box faces, gravity and wall directions to resolve ambiguity in camera orientation and refine the full pose.
This reframes sparse-map relocalization as verification of spatially extended video queries, moving discriminative support from stored appearance to temporal context and permitting a 100\,kB map of class-labelled boxes.
On RIO10 and ReplicaCAD, the all-frame localization success rate at 1\,m/10$^\circ$ is \textbf{73.5\,\%} and \textbf{61.1\,\%} under causal evaluation, rising to \textbf{90.6\,\%} and \textbf{74.8\,\%} with clip closure.
The evaluated per-frame scene coordinate regressors reach up to \textbf{47.6\,\%} and \textbf{49.8\,\%}, respectively, with maps of 12.6--42\,MB.
\ifanon\else Project page: \url{https://videoreloc.github.io}.\fi

%% file: sections/introduction.tex
\section{INTRODUCTION}
\label{sec:intro}

Camera relocalization estimates the six-degree-of-freedom camera pose in a previously mapped scene from a current observation~\cite{shotton2013scrf}.
Robots revisit indoor spaces to follow known routes and inspect places over time.
A map from an earlier visit provides the common spatial reference needed to compare observations and act at known locations.
Changes in lighting and furniture make aligning a later video to this reference difficult.
We therefore study long-term indoor video relocalization: given a sparse semantic scene-graph map, estimate a map-frame camera pose for every frame of the later video, as shown in Fig.~\ref{fig:teaser}.

Visual relocalizers rely on appearance recorded during mapping, so their reliability can fall as the scene changes.
This includes scene coordinate regression~\cite{shotton2013scrf,brachmann2023ace,jiang2025rscore,bruns2025aceg}, feature matching~\cite{sarlin2019hloc} and radiance-field localization~\cite{yen2021inerf}.
On RIO10~\cite{wald2020rio10}, ACE-G~\cite{bruns2025aceg} and R-SCoRe~\cite{jiang2025rscore} store tens of megabytes per room yet localize fewer than half of the re-scan frames within 1\,m and 10$^\circ$, as shown in Table~\ref{tab:main}.

Scene graphs instead encode object classes and geometry~\cite{salasmoreno2013slampp,nicholson2019quadricslam,armeni2019scenegraph,gu2024conceptgraphs}, reducing dependence on lighting but leaving a single-frame match ambiguous.
Moved furniture can give an incorrect correspondence, and a frame may see only a few objects whose class labels also occur elsewhere in the map.
Existing object-based relocalizers~\cite{zins2022oaslam,lee2026msgloc} must then choose among several map locations that explain the same observed object layout.
Object centers also omit surface orientation because a box can rotate in place without changing its center, and they provide no explicit vertical direction.
These limitations leave no well-established solution for estimating a complete video trajectory from a compact scene-graph map that stores no visual descriptors.

\method{} replaces mapping-visit appearance with a persistent map of class-labelled 3D boxes, about 100\,kB per room, and draws discriminative support from the incoming video.
Adaptive clips use odometry to turn partial frame observations into a spatially extended query, closing after sufficient object evidence and camera travel.

Within each clip query, separate modules address correspondence and orientation ambiguity.
To resolve correspondence ambiguity from repeated labels and moved objects, \emph{hypothesis-first registration} matches local and map object triangles to propose poses, ranks them by same-class center agreement across the entire clip, and verifies the winner by box-surface consistency.
To recover the orientation information absent from object centers, \emph{orientation-aware refinement} uses box faces, gravity and wall directions to update rotation and translation together.

Repeated layouts can still leave more than one plausible clip placement.
As the video arrives, tracking propagates the previous placement by camera motion, while weak or inconsistent evidence triggers another map search.
Motion-connected clips form a \emph{run}; the run-level decision uses their accumulated observations to retain or revise conflicting placements at clip closure.

On RIO10 and ReplicaCAD, the all-frame localization success rate at 1\,m/10$^\circ$ is 73.5\,\% and 61.1\,\% under causal evaluation, rising to 90.6\,\% and 74.8\,\% with clip closure, as shown in Table~\ref{tab:main}.
Controlled clip-length comparisons and module ablations measure the contributions of temporal aggregation, refinement and cross-clip consistency.
Together, the evidence identifies query-side temporal context as a central source of observability for long-term relocalization with sparse maps.

Our technical contributions are summarized as follows:
\begin{itemize}
  \item a \emph{video-clip relocalization paradigm} that constructs spatially extended queries from adaptive clips and uses a run-level decision to reconcile conflicting placements;
  \item \emph{hypothesis-first registration} that checks candidate poses against the clip's object layout to reject geometrically inconsistent matches;
  \item \emph{orientation-aware refinement} that uses oriented box-face geometry to update rotation and translation jointly, with gravity constraining camera tilt and wall directions constraining camera yaw.
\end{itemize}

%% file: sections/related_work.tex
\section{RELATED WORK}
\label{sec:related}

\begin{figure*}[t]
  \centering
  \includegraphics[width=0.92\textwidth]{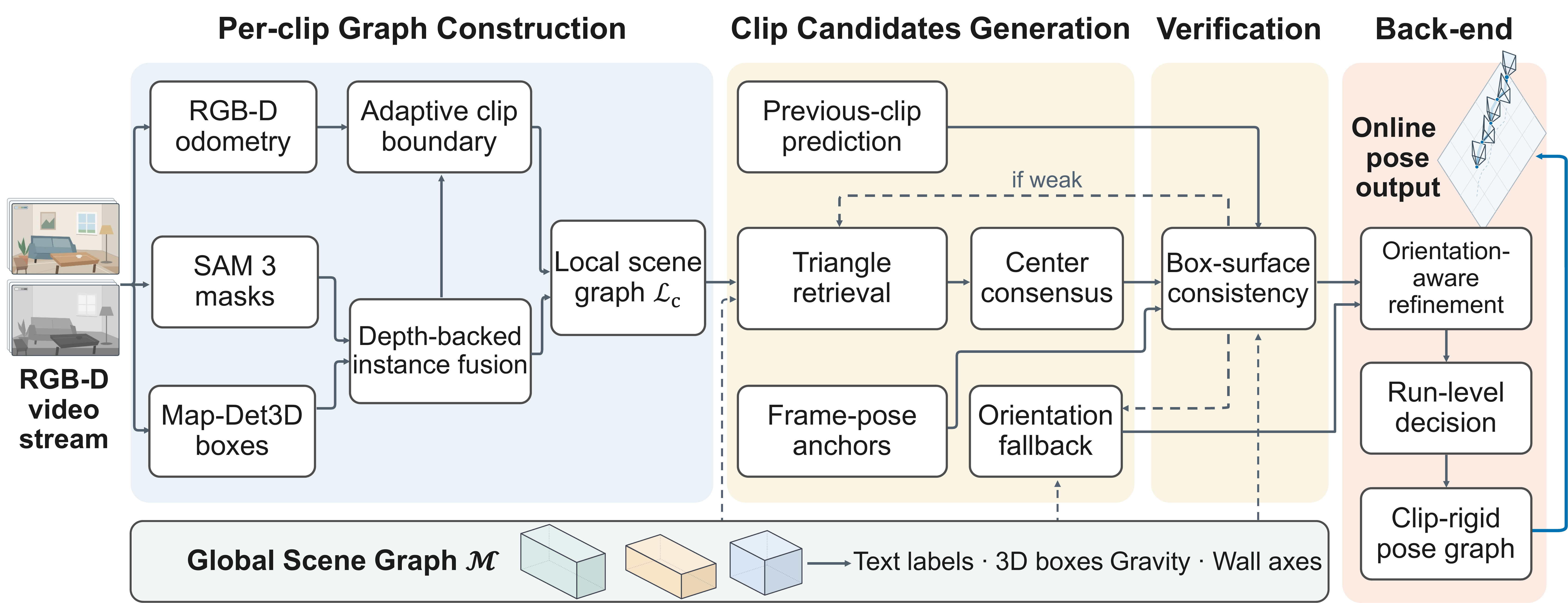}
  \caption{\textbf{Overview of \method{}.} Per-clip graph construction produces $\mathcal{L}_c$ from RGB-D odometry and depth-backed instance fusion. Clip candidates generation combines previous-clip prediction, triangle retrieval, frame-pose anchors and orientation fallback; box-surface consistency verifies them. The back-end applies orientation-aware refinement, the run-level decision and the clip-rigid pose graph; final frame poses satisfy $T_i=S_cV_i$.}
  \label{fig:pipeline}
\end{figure*}

\textbf{Appearance-based relocalization.}
Appearance-based relocalizers regress scene coordinates~\cite{shotton2013scrf,brachmann2023ace,jiang2025rscore,bruns2025aceg}, match point-cloud descriptors~\cite{sarlin2019hloc}, or align radiance-field renderings~\cite{yen2021inerf}; their dependence on mapping-visit appearance can reduce reliability after change~\cite{wald2020rio10}.
\method{} instead stores object classes and geometry.

\textbf{Object-level relocalization and scene graphs.}
Object maps represent landmarks with models~\cite{salasmoreno2013slampp}, dual quadrics~\cite{nicholson2019quadricslam}, or scene graphs that add spatial organization~\cite{armeni2019scenegraph,gu2024conceptgraphs}.
Hydra~\cite{hughes2022hydra} and S-Graphs~\cite{bavle2023sgraphs} build hierarchical maps; SGAligner~\cite{sarkar2023sgaligner}, SG-Reg~\cite{liu2025sgreg} and ROMAN~\cite{peterson2024roman} align submaps from graph, semantic or shape cues.

OA-SLAM~\cite{zins2022oaslam}, MSG-Loc~\cite{lee2026msgloc} and GOReloc~\cite{wang2024goreloc} associate frame observations with mapped objects; GOReloc uses RANSAC-style verification of graph associations.
OpenReLoc~\cite{cui2026openreloc} combines open-vocabulary 2D--3D object matching with shape-guided image alignment, while SceneGraphLoc~\cite{miao2024scenegraphloc} predicts coarse image-to-graph locations.
These methods process individual queries; \method{} accumulates adaptive video clips against a class-and-box map and returns every frame pose.

\textbf{Triangle-based matching and registration.}
Sparse geometric layouts support global registration without dense appearance matching.
STD~\cite{yuan2023std} indexes keypoint triangles by side length.
Outram~\cite{yin2024outram} triangulates a LiDAR cluster map, pools correspondences, selects a maximum clique, and robustly fits a transform.
Liang and Pei~\cite{liang2026geometric} use triangle constraints to seed and incrementally expand geometrically consistent node correspondences between two scene graphs.
\method{} instead turns each triangle match into a clip-pose hypothesis, ranks it with all same-class clip centers, and compares its box-surface support with an anchor-derived candidate.
Section~\ref{sec:ablations} compares these hypothesis-first and correspondence-first routes on identical local scene graphs.

%% file: sections/method.tex
\section{METHOD}
\label{sec:method}

\subsection{Overview}
\label{sec:overview}
\label{sec:problem}
Given a sparse semantic scene-graph map from an earlier visit, the task is to estimate a map-frame camera pose $T_i$ for the $i$-th frame of a later RGB-D video.
A single frame often exposes only a partial object set that can recur elsewhere, whereas camera motion connects complementary views.
We therefore treat each odometry-connected frame sequence, or clip $c$, as one spatial query: RGB-D visual odometry supplies frame pose $V_i$ in clip coordinates, and \method{} estimates one clip-to-map transform $S_c$ without altering the relative geometry.
The back-end value after frame $b$ is $S_c^{[b]}$, with $[b]$ omitted when the cutoff is clear (Fig.~\ref{fig:pipeline}); thus
\begin{equation}
  T_i = S_c\,V_i, \qquad i \in c .
  \label{eq:clip}
\end{equation}
The pipeline has four stages: per-clip graph construction fuses $\mathcal{L}_c$, clip candidates generation proposes placements, verification selects one, and the back-end refines it and reconciles clips across the run.

\subsection{Stored global map}
\label{sec:map}
Appearance changes weaken visual correspondence, while dense maps are costly to store.
We therefore store semantic identity and coarse 3D extent as class-labelled boxes, deriving searchable relations without visual descriptors.
The global map $\mathcal{M}$ contains gravity-aligned oriented boxes $m_j$ with plain-text class names $\ell_j$ for objects, walls, floors and ceilings.
Following HOV-SG~\cite{hovsg}, map construction combines class-agnostic 3D segmentation~\cite{huang2024segment3d} with open-vocabulary labelling~\cite{sam3}.
At load time, nearby object pairs form center-distance edges, and non-degenerate triples form an index keyed by vertex classes and edge lengths; both are derived rather than serialized.
Gravity alignment supplies map vertical $+z$, while an extent-weighted circular median of the wall boxes' longest horizontal axes, modulo $90^\circ$, supplies the Manhattan axes.
RIO10 room maps occupy 45--164\,kB.

\subsection{Per-clip graph construction}
\label{sec:localgraph}
A frame observes only visible object surfaces, so its object set is incomplete and its estimated centers are view-dependent.
We therefore use motion to place class-labelled observations in a common clip frame and fuse repeated views before registration.
Open3D's geometry-based \mbox{RGB-D} odometry~\cite{newcombe2011kinectfusion,zhou2018open3d} chains consecutive motions into within-clip poses $V_i$.
At a regular stride and at clip end, SAM 3~\cite{sam3} segments sampled RGB video frames using the map's class names.
Depth-consistent mask pixels are back-projected; their centroid gives the observation center, and farthest-point sampling retains up to twelve visible-surface samples for box matching.
To reduce center bias, an unused Map-Det3D~\cite{mapdet3d} box replaces the mask center when its projected box has the highest 2D IoU with the SAM 3 detection box and passes the depth check.
After $V_i$ maps each observation into clip coordinates, connected components of nearby same-class observations form fused nodes.
Each node $k\in\mathcal{K}_c$ stores class $\ell_k$, mean center, and accumulated surface samples $P_k$.

Fixed windows can end before enough objects are observed or extend beyond reliable odometry.
Adaptive clips therefore close once fused evidence and camera travel are sufficient, but end earlier at capture gaps or inconsistent odometry.

\subsection{Clip candidates generation and verification}
\label{sec:registration}
Repeated labels make a complete local-to-map assignment ambiguous, and one early mismatch can corrupt the pose.
We instead generate poses from small class-compatible constellations and let all eligible local instances select the globally coherent hypothesis.
Visible room structures do not share their mapped boxes' centers, so walls, floors and ceilings are excluded from center proposals but retained for orientation (Sec.~\ref{sec:cues}) and surface verification.
Let $\mathcal{U}_c$ contain the remaining local instances whose classes occur in the map.
The indexed route bounds this set by prioritizing non-movable classes and then $|P_k|$; its centers form $N$ non-degenerate triplets $\{\mathcal{T}_n^{\mathrm{loc}}\}_{n=1}^{N}$ that query map triplets with compatible classes and edge lengths.
Matches with fewer movable vertices and smaller edge mismatch rank first, and we retain up to 500.
The complementary random route draws three distinct members of the full $\mathcal{U}_c$ and assigns them to distinct same-class map nodes, favoring compatible neighboring classes and distances.
Together, the two routes produce $H$ minimal assignments $\{\mathcal{P}_h\}_{h=1}^{H}$.
Each $\mathcal{P}_h=\{(x_{h,a},y_{h,a})\}_{a=1}^{3}$ pairs three same-class local and map centers; rigid least squares gives $S_h=(R_h,t_h)$:
\begin{equation*}
  (R_h,t_h)\in
  \operatorname*{arg\,min}_{R\in SO(3),\,t\in\mathbb{R}^3}
  \sum_{a=1}^{3}\left\|R x_{h,a}+t-y_{h,a}\right\|^2.
\end{equation*}
For each $S_h$, we greedily form non-repeating same-class pairs $A_h$ from all $\mathcal{U}_c$ centers within 0.5\,m; $e_h$ is their root-mean-square distance ($+\infty$ if empty).
We select $h^\star=\arg\max_h\bigl(|A_h|,-e_h\bigr)_{\mathrm{lex}}$: most matches first, lowest error on ties.
Refitting $S_{h^\star}$ using all center pairs in $A_{h^\star}$ yields the graph-derived clip-to-map candidate $S_{\mathrm{g}}$.

A sparse or repetitive local graph can make $S_{\mathrm{g}}$ unavailable or weak even when individual frames admit reliable fits.
We therefore turn independently fitted frames into temporal pose anchors whose agreement can supply a placement or veto a conflicting graph hypothesis.
The route runs when $S_{\mathrm{g}}$ is absent, has fewer than six center matches, or fewer than $60\,\%$ of instances pass the 0.5\,m box-surface test.
Per-frame RANSAC samples three class-compatible observation/map-center pairs, ranks poses by 0.3\,m box-surface inliers, and refits the winner.
A fit with at least six inliers and at most 0.5\,m root-mean-square error gives a map-frame anchor $W_i$ and proposal $\widehat S_i=W_iV_i^{-1}$.
RANSAC selects the $\widehat S_i$ supported by the most anchors $j$, requiring the translation distance plus rotation angle in radians between $\widehat S_iV_j$ and $W_j$ to be below 0.6.
A rotation average and corresponding mean translation over at least three supporting anchors yield the alternative clip-to-map candidate $S_{\mathrm{a}}$.
The anchors also reject $S_{\mathrm{g}}$ when the median translation or rotation disagreement between $S_{\mathrm{g}}V_i$ and $W_i$ exceeds 1.5\,m or $40^\circ$.

Center consensus can select the wrong repeated layout, so both routes need a common reliability test.
We therefore verify whether each candidate places fused object and structural surfaces in or near same-class solid boxes.
For each available candidate $S=(R,t)\in\{S_{\mathrm{g}},S_{\mathrm{a}}\}$, let $S(p)=Rp+t$ map a surface-point sample $p$ from clip to map coordinates.
For fused instance $k$ and map node $m_j$, $\rho_{kj}(S)$ is the root-mean-square point-to-box distance of $\{S(p):p\in P_k\}$ to $m_j$'s solid box; interior points contribute zero.
With $\tau=0.5$\,m, the score is
\begin{equation}
  \kappa(S)=\frac{1}{|\mathcal{K}_c|}\sum_{k\in\mathcal{K}_c}
  \mathds{1}_{\left\{\min_{j:\,\ell_j=\ell_k}\rho_{kj}(S)<\tau\right\}},
  \label{eq:cons}
\end{equation}
Here $|\mathcal{K}_c|$ counts the fused instances in clip $c$, $\mathds{1}_{\{\cdot\}}$ is the indicator, and the minimum is $+\infty$ without a class match.
An indicator value of one assigns $k$ to its minimizing same-class box.
Candidates rank by $\kappa(S)$; ties use the smaller median of $\min_{j:\,\ell_j=\ell_k}\rho_{kj}(S)$ over $k\in\mathcal{K}_c$ with a same-class map node.
The winner is denoted $S_{c,0}$, the input to the first orientation-aware refinement solve.
The aggregate score $\kappa(S)$ also controls tracking, while instances passing its per-instance test are used for surface refinement.

When graph correspondences and pose anchors fail to place a clip, gravity and wall directions can still constrain its rotation.
This orientation-based fallback aligns the clip's gravity direction with the map's vertical axis, fixing camera tilt, and aligns the observed wall directions with the map's Manhattan axes to obtain a reference camera yaw $\psi_0$.
Because Manhattan directions determine camera yaw only modulo $90^\circ$, we enumerate four rotations $R_k$ with yaw $\psi_k=\psi_0+90^\circ k$, $k\in\{0,1,2,3\}$, while keeping camera tilt fixed.
For each $R_k$, every same-class local/map-center pair $(x,y)$ proposes $t=y-R_kx$; the proposal with the most one-to-one inliers within 0.5\,m is replaced by the mean offset over those inliers.
When an odometry prediction is available, only rotations within $45^\circ$ of its yaw are retained.
The box-surface score restricted to object instances selects among the survivors.
When the fallback is used, its winner likewise defines $S_{c,0}$.

\begin{table*}[!t]
  \caption{Comparison of recall and pose errors across relocalization methods and evaluation variants on RIO10 and ReplicaCAD.}
  \label{tab:main}
  \centering
  \small
  \setlength{\tabcolsep}{2.2pt}
  \begin{tabular}{@{}lccccccc@{}}
    \toprule
    & & \multicolumn{3}{c}{RIO10, 10 re-scans, 34{,}415 frames} & \multicolumn{3}{c}{ReplicaCAD, 18 tours, 72{,}000 frames} \\
    \cmidrule(lr){3-5}\cmidrule(l){6-8}
    Method & Map size $\downarrow$ & Pos./rot. $\downarrow$ & \recall{} $\uparrow$ & 0.25\,m/5$^\circ$ $\uparrow$ & Pos./rot. $\downarrow$ & \recall{} $\uparrow$ & 0.25\,m/5$^\circ$ $\uparrow$ \\
    \midrule
    HLoc~\cite{sarlin2019hloc}            & 1--2\,GB & \phantom{0}12\,cm / \phantom{0}3.9$^\circ$ & 49.8 & 41.3 & \phantom{00}\underline{4\,cm} / \phantom{0}\underline{0.6$^\circ$} & 48.4 & 44.5 \\
    ACE-G~\cite{bruns2025aceg}          & 12.6\,MB & \phantom{0}36\,cm / 11.7$^\circ$ & 45.5 & 25.7 & \phantom{0}52\,cm / \phantom{0}8.2$^\circ$ & 49.8 & 37.1 \\
    R-SCoRe~\cite{jiang2025rscore}      & 42\,MB   & \phantom{0}40\,cm / 12.1$^\circ$ & 47.6 & 38.5 & \phantom{0}72\,cm / \phantom{0}9.4$^\circ$ & 48.6 & 42.5 \\
    MSG-Loc~\cite{lee2026msgloc}            & \underline{1.5\,MB} & 161\,cm / 49.3$^\circ$ & \phantom{0}1.2 & \phantom{0}0.1 & 265\,cm / 27.7$^\circ$ & 11.3 & \phantom{0}1.3 \\
    \midrule
    FPFH--ICP$^\dagger$                   & 6--13\,MB & \phantom{0}11\,cm / \phantom{0}3.4$^\circ$ & 22.4 & 21.7 & \phantom{00}\textbf{1\,cm} / \phantom{0}\textbf{0.2$^\circ$} & 10.0 & \phantom{0}9.9 \\
    ACE-G$^\dagger$                       & 12.6\,MB & \phantom{0}\underline{10\,cm} / \phantom{0}3.6$^\circ$ & 79.8 & \underline{66.2} & \phantom{0}14\,cm / \phantom{0}1.8$^\circ$ & \underline{67.0} & \underline{59.4} \\
    R-SCoRe$^\dagger$                     & 42\,MB   & \phantom{00}\textbf{9\,cm} / \phantom{0}\textbf{3.0$^\circ$} & \underline{85.8} & \textbf{71.0} & \phantom{0}11\,cm / \phantom{0}1.7$^\circ$ & 66.7 & \textbf{59.8} \\
    \midrule
    \method{} (ours), causal            & \textbf{95\,kB} & \phantom{0}20\,cm / \phantom{0}4.3$^\circ$ & 73.5 & 44.0 & \phantom{0}27\,cm / \phantom{0}3.0$^\circ$ & 61.1 & 40.4 \\
    \method{} (ours), with clip closure & \textbf{95\,kB} & \phantom{0}18\,cm / \phantom{0}\underline{3.3$^\circ$} & \textbf{90.6} & 64.1 & \phantom{0}22\,cm / \phantom{0}2.1$^\circ$ & \textbf{74.8} & 53.0 \\
    \bottomrule
  \end{tabular}
  \par\smallskip
  \begin{minipage}{\textwidth}
  \footnotesize
  \textit{Protocol.} Recall (\%) includes all frames, with missing poses counted as failures. Causal evaluation uses frames up to the query frame. Evaluation with clip closure also uses the remaining frames of that clip. Bold and underlined values mark the best and second-best results per metric, respectively; median position and rotation errors are ranked separately, and ties share rank. $^\dagger$ marks clip-rigid controls; for methods that normally operate per frame, it denotes their clip-based variants. These controls use the same clip partition and within-clip odometry as \method{}, but localize each clip independently without cross-clip tracking, run-level decisions, or pose-graph optimization.

  \textit{Map size and error statistics.} The map size column reports the size of the map constructed by each method on RIO10. ReplicaCAD maps occupy 134\,kB for \method{}, 1.96\,MB for MSG-Loc and 18.1\,MB for FPFH--ICP. Median position and rotation errors use only returned poses. Pose coverage (RIO10 / ReplicaCAD) is 100/100\,\% for regressors, 75/68\,\% for HLoc, 29/12\,\% for FPFH--ICP, 21/61\,\% for MSG-Loc, 94/97\,\% for \method{} under causal evaluation and 100/99\,\% with clip closure. FPFH--ICP errors summarize scene medians on RIO10 and a separate replay on ReplicaCAD.
  \end{minipage}
\end{table*}

\subsection{Orientation cues and refinement}
\label{sec:cues}
Object-center geometry can place a clip while leaving its orientation ambiguous; gravity and the room's persistent Manhattan structure provide the missing tilt and yaw references, respectively.
In the gravity-aligned map frame, $+z$ is vertical and $u$ is the clip-frame vertical.
For a candidate rotation $R$, gravity aligns $Ru$ with $+z$, while Manhattan walls constrain the remaining yaw about $+z$.

\textbf{Gravity.}
Under an upright-object assumption, the detector-confidence-weighted mean of detected box up-axes estimates $u$ in clip coordinates.
When box cues are unavailable, floor and ceiling plane normals together with the direction orthogonal to wall normals provide $u$; the mean camera-up direction from $V_i$ is the final fallback.
A candidate $S=(R,t)$ passes the gravity gate only if $\angle(Ru,+z)\leq30^\circ$, or $60^\circ$ for the camera-up fallback.
Accepted transforms are then levelled about the clip centroid.

\textbf{Manhattan yaw.}
The candidate rotation maps each clip-frame wall-plane normal into the global map frame, where its horizontal direction should coincide with the nearest map Manhattan axis.
We measure this angular discrepancy modulo $90^\circ$ to account for normal sign and the two orthogonal room axes.
A yaw correction is accepted only when at least five normals form a concentrated consensus (circular median absolute deviation no greater than $12^\circ$); their circular median supplies the correction.
Otherwise, yaw remains unchanged.

\textbf{Orientation-aware refinement.}
An orientation correction alone can leave positional error from an earlier fit.
We therefore refine rotation and translation together after each accepted gravity or yaw correction.
The core idea is to use each assigned same-class map box as a one-sided geometric constraint: exterior surface samples correct the pose, while samples already explained by the box volume incur no penalty.

For each solve, $S_{c,0}$ denotes its input: the selected candidate initially and the cue-corrected transform after a cue update.
At each iteration, $\mathcal{I}\subseteq\mathcal{K}_c$ contains the instances passing the box-surface consistency test under the current $S$, with the assignments defined above.
The objective jointly minimizes one-sided surface-point-to-box error, camera tilt error and departure from the input transform:
\begin{equation}
  \begin{split}
  E(S)={}&\sum_{k\in\mathcal{I}}\sum_{p\in P_k}
  \frac{\mathds{1}_{\left\{S(p)\notin\mathcal{B}_k\right\}}\,\|S(p)-q_p\|_2^2}{|P_k|\,\sigma_f^2}\\
  &+\frac{\|r_g(S,u)\|_2^2}{\sigma_g^2}
  +\big\|r_0(S,S_{c,0})\big\|^2_{\Sigma_0^{-1}} .
  \end{split}
  \label{eq:refine}
\end{equation}
Here $\mathcal{B}_k$ is the assigned solid box; $q_p$ is its closest point to $S(p)$; $|P_k|$ is the number of point samples of instance $k$; and $\|\cdot\|_2$ and $\|\cdot\|_{\Sigma_0^{-1}}^2$ denote the Euclidean and squared matrix-weighted norms.
The indicator suppresses interior samples; for exterior ones, the shortest box displacement corrects normal error without imposing a tangential point match.
Division by $|P_k|$ gives every instance equal surface weight.

The gravity residual $r_g(S,u)$ measures camera tilt error without constraining yaw or translation.
The prior residual $r_0(S,S_{c,0})$ measures translation and rotation changes from $S_{c,0}$, and $\Sigma_0$ weights these changes separately.
The scales $\sigma_f$ and $\sigma_g$ balance the surface-matching and tilt residuals.
The box assignments, closest-point projections and set $\mathcal{I}$ are updated as the estimate changes.
The output after the last refinement solve is denoted $\widetilde S_c$, the refined front-end placement of clip $c$.

\subsection{Back-end online operation}
\label{sec:online}
For each new clip, online operation proceeds in the order of tracking, the run-level decision and pose-graph reconciliation.

\textbf{Tracking across clips.}
Tracking avoids jumps between repeated layouts and blind drift accumulation by treating inter-clip motion as a proposal verified against the map.
For the boundary frames $p\in c$ and $q\in c+1$, the \emph{boundary odometry} $O_c$ is the RGB-D odometry transform from $p$ to $q$ and predicts $S_{c+1}^{\mathrm{pred}}=\widetilde S_cV_pO_cV_q^{-1}$.
Surface samples from clip $c+1$ validate this prediction against same-class boxes; sufficient support accepts and refines it, whereas failure reruns the localization and orientation stages in Secs.~\ref{sec:registration} and~\ref{sec:cues}.

\textbf{Run-level decision.}
A clip can fit the wrong copy of a repeated layout, making its local score ambiguous.
We therefore judge its placement with evidence beyond that clip while preserving independently supported alternatives.
A \emph{run} is a maximal sequence of clips connected by unbroken boundary odometry.
With $C_r=I$ for the run's first clip, $C_{c+1}=C_cV_pO_cV_q^{-1}$ maps each subsequent clip into the common run frame.
Each front-end placement proposes a run-to-map hypothesis $G_c=\widetilde S_cC_c^{-1}$, and any $G$ predicts clip $d$'s placement as $GC_d$.
We score the $G_c$ and gravity/wall-derived orientation variants over all observed clips by summing the per-instance test in Eq.~\eqref{eq:cons}.
The maximizer $G^\star$ is the reference run-to-map hypothesis against which conflicting clip placements are judged.
A \emph{placement-consistent sub-run} is a maximal consecutive part of the run whose neighboring $G_c$ agree within 0.5\,m and $10^\circ$.
The source sub-run of $G^\star$ seeds the trusted set; for a cue-derived winner, any sub-run of at least two clips can seed it.
Other sub-runs retain their placements only when their accumulated support exceeds propagation from the nearest trusted sub-run; otherwise, their placements are replaced by the propagated ones.
Replacements are box-refined under a 0.5\,m/$5^\circ$ prior, and boundary edges inconsistent with the selected placements are removed.

\textbf{Clip-rigid pose graph.}
Accepted placements and boundary odometry can disagree, so treating them independently would create jumps at clip boundaries.
We therefore optimize one rigid node $S_c$ per clip, reconciling the discrepancy without deforming within-clip motion.
Relative edges enforce $S_c^{-1}S_{c+1}\approx V_pO_cV_q^{-1}$; absolute edges anchor $S_c$ to $\widetilde S_c$ or its run-level revision.
At clip closure, Gauss--Seidel relaxation with chordal rotation averaging~\cite{hartley2013rotation} and a Geman--McClure kernel~\cite{black1996robust} updates the graph; inconsistent object-graph constraints are demoted from weight 20 to 1.
Each update moves whole clips rigidly, preserving their internal odometry trajectories.
Gravity and wall-direction corrections are reapplied, and the corrected $S_c$ gives the pose output in Eq.~\eqref{eq:clip}.

%% file: sections/experiments.tex
\input{tables/replicacad_factors}

\begin{table*}[!t]
  \begin{minipage}[t]{0.265\textwidth}
    \caption{Effect of the localization unit on RIO10.}
    \label{tab:unit}
    \centering
    \footnotesize
    \begin{tabular}{@{}lcc@{}}
      \toprule
      Unit & 1\,m/10$^\circ$ & 0.25\,m/5$^\circ$ \\
      \midrule
      one frame        & \phantom{0}2.6 & \phantom{0}0.4 \\
      30-frame clip    & 19.1 & \phantom{0}9.3 \\
      100-frame clip   & 39.3 & 23.7 \\
      250-frame clip   & 52.8 & 35.4 \\
      \bottomrule
    \end{tabular}
    \par\smallskip\raggedright
    Each unit is registered independently, without tracking or a pose graph. Recall (\%, higher is better) uses all 34{,}415 frames and is measured when the unit closes.
  \end{minipage}\hfill
  \begin{minipage}[t]{0.705\textwidth}
    \caption{Effects of refinement, run-level decisions and adaptive clips.}
    \label{tab:ablation}
    \centering
    \footnotesize
    \setlength{\tabcolsep}{2.6pt}
    \begin{tabular}{@{}ccccccccccc@{}}
      \toprule
      \multicolumn{3}{c}{Mechanism} & \multicolumn{4}{c}{RIO10} & \multicolumn{4}{c}{ReplicaCAD tours} \\
      \cmidrule(r){1-3}\cmidrule(lr){4-7}\cmidrule(l){8-11}
      \multirow[c]{2}{*}{orientation-aware} & \multirow[c]{2}{*}{run-level} & \multirow[c]{2}{*}{adaptive} & \multicolumn{2}{c}{with clip closure} & \multicolumn{2}{c}{causal} & \multicolumn{2}{c}{with clip closure} & \multicolumn{2}{c}{causal} \\
      \cmidrule(lr){4-5}\cmidrule(lr){6-7}\cmidrule(lr){8-9}\cmidrule(l){10-11}
      & & & coarse & fine & coarse & fine & coarse & fine & coarse & fine \\
      \midrule
                 &            &            & 76.4 & 39.3 & 70.8 & 33.4 & 64.9 & 38.6 & 56.3 & 33.6 \\
      \checkmark &            &            & 78.5 & 44.4 & 72.5 & 38.0 & 70.4 & 45.1 & \underline{62.0} & \underline{41.3} \\
      \checkmark & \checkmark &            & 81.3 & 43.9 & 70.8 & 36.2 & \textbf{78.9} & \underline{51.6} & \textbf{69.4} & \textbf{47.1} \\
      \checkmark &            & \checkmark & \textbf{91.0} & \underline{61.3} & \underline{73.0} & \underline{41.8} & 56.3 & 39.3 & 45.4 & 29.2 \\
      \checkmark & \checkmark & \checkmark & \underline{90.6} & \textbf{64.1} & \textbf{73.5} & \textbf{44.0} & \underline{74.8} & \textbf{53.0} & 61.1 & 40.4 \\
      \bottomrule
    \end{tabular}
    \par\smallskip\raggedright
    Recall (\%, higher is better) includes all frames. Coarse: 1\,m/10$^\circ$. Fine: 0.25\,m/5$^\circ$. Checkmarks indicate enabled components; empty cells indicate point-to-point refinement, no run-level decision and fixed 100-frame clips, respectively. Tracking and the pose graph remain enabled. Bold/underline: first/second in each column.
  \end{minipage}
\end{table*}

\begin{figure*}[!t]
  \centering
  \includegraphics[width=\textwidth]{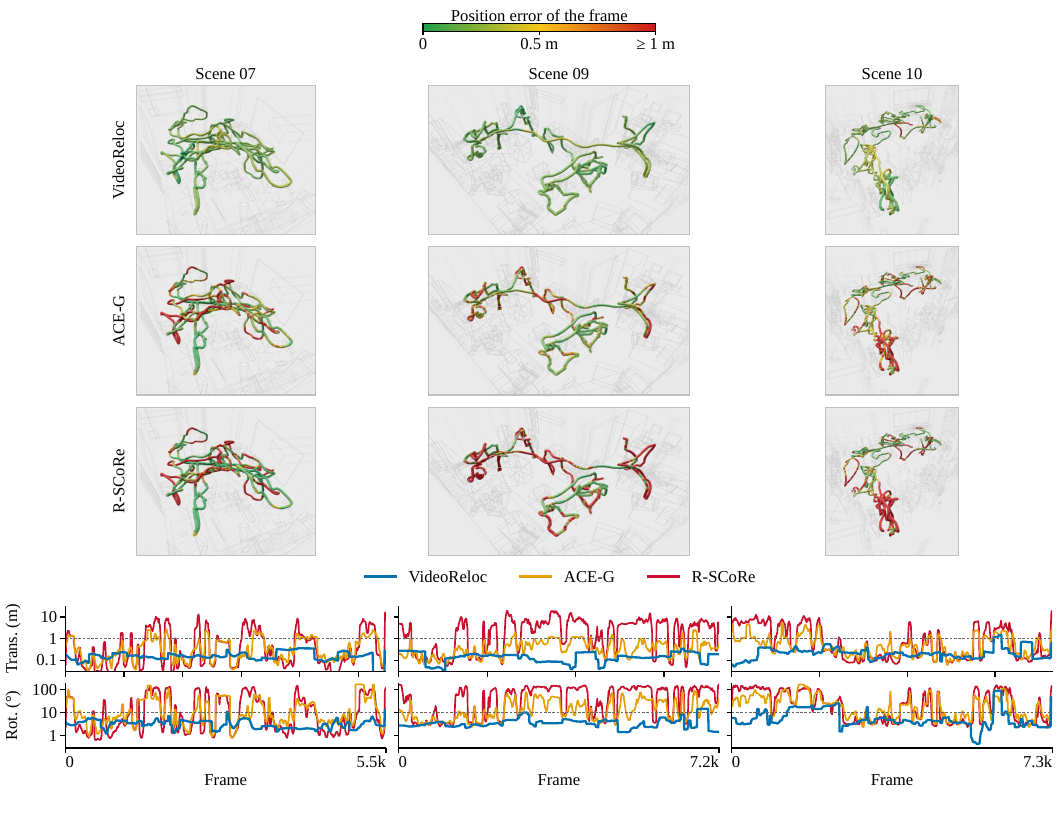}
  \caption{\textbf{RIO10 qualitative results.} Grey boxes are map objects; trajectory colour continuously encodes position error (green: 0\,m; yellow: 0.5\,m; red: $\geq$1\,m). \method{} uses estimates with clip closure. Lower plots show log-scale rolling median translation/rotation errors; dashed lines mark 1\,m/10$^\circ$.}
  \label{fig:qual}
\end{figure*}

\section{EXPERIMENTS}
\label{sec:experiments}

\subsection{Setup}
\label{sec:setup}
RIO10~\cite{wald2020rio10} contains ten rooms scanned weeks apart with furniture and lighting changes; maps use the first visit, and evaluation includes all 34{,}415 re-scan frames.
We construct a controlled ReplicaCAD~\cite{szot2021habitat2} evaluation from six authored layouts of one apartment.
Using \texttt{apt\_0} as the reference, we evaluate five variants that retain the shell but move 68--76 objects by more than 0.1\,m (59--68 by more than 0.5\,m; moved-object median 1.9--3.5\,m), add 8--10, and remove 1--10.
In Blender, we render the same 4{,}000-frame path in each layout under daylight with white fills, warm evening lights, and two warm night lamps, yielding a $6\times3$ RGB-D design that separates illumination from rearrangement effects.
We use a separate 6{,}000-frame cross-hatch tour of \texttt{apt\_0} under daylight to build the map for all methods.

\textbf{Implementation details.}
Clips $c$ begin at 100 frames, grow by 50 to at most 600, and close once $|\mathcal{K}_c|\geq20$ and the camera path reaches 2\,m.
Object observations use every fifth frame and the final frame.
A mask--box pair requires 2D IoU $>0.4$ and center-depth difference $\leq0.4z+0.5$\,m, where $z$ is the mask-center depth.
At most 48 local instances form the $N$ query triplets $\{\mathcal{T}_n^{\mathrm{loc}}\}_{n=1}^{N}$; compatible map triangles have 0.2--8\,m edges within 0.3\,m of the query.
The $H$ assignments $\{\mathcal{P}_h\}_{h=1}^{H}$ combine at most 500 indexed matches with 2{,}000 random draws of three distinct members of $\mathcal{U}_c$ to distinct same-class map nodes.
We rank up to 25 map-frame anchors $W_i$ having at least six matches and at most 0.5\,m root-mean-square error by match count then error; three RANSAC-consistent anchors are required to estimate $S_{\mathrm{a}}$.

\textbf{Evaluation protocol.}
For evaluation, two evidence cutoffs freeze $T_i$; later clips cannot update it.
\emph{Causal} evaluation uses observations only through the $i$-th frame.
It propagates the preceding clip's saved placement with odometry; at run starts, it uses per-frame localization when available.
Evaluation \emph{with clip closure} waits for clip $c$ containing that frame to close.
At closure, the run-level decision and incremental clip-rigid pose-graph update use the clip's refined placement to produce $S_c$; Eq.~\eqref{eq:clip} gives the frozen pose $T_i=S_cV_i$.

Recall counts a frame as successful only when neither position nor rotation error exceeds the stated thresholds.
The primary threshold is 1\,m/10$^\circ$, and the finer threshold is 0.25\,m/5$^\circ$.
Frames without an estimate count as failures.
Median errors use frames with an estimate.

\textbf{Baselines.}
HLoc~\cite{sarlin2019hloc} uses NetVLAD, SuperPoint and LightGlue.
ACE-G~\cite{bruns2025aceg} and R-SCoRe~\cite{jiang2025rscore} use their official implementations and the same mapping data as \method{}.

Our implementation reproduces MSG-Loc's published accuracy on the authors' sequence~\cite{lee2026msgloc}.
On RIO10, we then evaluate MSG-Loc with either its original detector or the same RGB-D video and SAM 3 detections as \method{}.
In both settings, MSG-Loc builds its own quadric map from the mapping video.

The $^\dagger$ controls use the clip-rigid protocol defined in Table~\ref{tab:main}: \method{}'s clip partition and within-clip odometry, independent localization at closure, and no cross-clip backend.
FPFH--ICP$^\dagger$ registers each clip's fused depth to a 5\,cm point-cloud map with FPFH descriptors, RANSAC and ICP~\cite{rusu2009fpfh,fischler1981ransac,besl1992icp}, providing a dense geometric control without object semantics.
ACE-G$^\dagger$ and R-SCoRe$^\dagger$ robustly fit one clip-to-map transform from frame poses and odometry; without a consensus, they retain the frame estimates.
RIO10 results are averaged over three sampling seeds.

\subsection{Long-term relocalization on RIO10}
\label{sec:rio10}
On RIO10, \method{} reaches $73.5\,\%$ recall at 1\,m/10$^\circ$ under causal evaluation and $90.6\,\%$ with clip closure, versus 45.5--49.8\,\% for the appearance-based per-frame baselines, with a 95\,kB mean map rather than 12.6--42\,MB, as shown in Table~\ref{tab:main}.

In the three scenes shown in Fig.~\ref{fig:qual}, \method{} has steadier errors and a higher success rate than ACE-G and R-SCoRe.

\subsection{Illumination and furniture changes in ReplicaCAD}
\label{sec:replicacad}

With a 134\,kB map of 294 boxes, \method{} reaches $61.1\,\%$ recall under causal evaluation and $74.8\,\%$ with clip closure, as shown in Table~\ref{tab:main}.

Day-to-night recall changes by $-1.0$ point under causal evaluation and $+0.1$ with clip closure, while the appearance-based per-frame baselines lose 6.3--17.7 points, showing the lower illumination sensitivity of object-class matching in Table~\ref{tab:replicacad_factors}.

Under furniture rearrangement, \method{} retains $71\,\%$ of unchanged-layout recall under causal evaluation and $81\,\%$ with clip closure, versus 56--58\,\% for the appearance-based per-frame baselines; FPFH--ICP$^\dagger$ retains $8\,\%$ and returns poses for roughly $12\,\%$ of frames, as shown in Tables~\ref{tab:replicacad_factors} and~\ref{tab:main}.

The ACE-G$^\dagger$ clip-rigid control reduces the day-to-night and rearrangement losses to 2.4 and 26.8 points, respectively.
Fitting one transform across a clip can suppress isolated frame errors.
Under rearrangement, ACE-G$^\dagger$ and R-SCoRe$^\dagger$ remain about ten points below \method{} at the same cutoff.

\subsection{Effect of the localization unit}
\label{sec:msgloc}
Without tracking or the pose graph, recall rises from $2.6\,\%$ for one frame to $52.8\,\%$ for a 250-frame clip under the same map, detector, registration and refinement, as shown in Table~\ref{tab:unit}.
The intermediate clip lengths show a consistent increase.
In our single-frame test, $85\,\%$ of frames contain at least three mappable objects, yet only $36\,\%$ produce a fit; $7\,\%$ of those fits are correct at 1\,m/10$^\circ$, with a $65^\circ$ median rotation error.

MSG-Loc's detector yields 0.7 objects/frame and $0.7\,\%$ pose coverage; matched-input SAM 3 yields $21\,\%$ coverage, $1.2\,\%$ recall, and a $49.3^\circ$ median rotation error.

\subsection{Ablation studies}
\label{sec:ablations}
Table~\ref{tab:ablation} shows that orientation-aware refinement improves fine recall, while the run-level decision mainly benefits ReplicaCAD and recovers losses when adaptive clips span repeated layouts within the 2\,m travel limit.

With anchors, refinement, tracking, and the run-level decision disabled, the hypothesis-first/correspondence-first rates on identical local graphs are $35.2/14.4\,\%$ on RIO10 and $36.2/22.5\,\%$ on ReplicaCAD (median-frame 1\,m/10$^\circ$), confirming more reliable clip-wide verification.

\subsection{Observation budget and computational cost}
\label{sec:limitations}
Delayed-output replay freezes $T_i=S_c^{[b]}V_i$ at the latest processed frame $b$ satisfying $0\leq b-i\leq B$, including across clip boundaries.
Recall rises monotonically from $73.5\,\%$ at $B=0$ to $79.7\,\%$ at $B=120$ on RIO10, and from $61.1\,\%$ to $77.7\,\%$ at $B=360$ on ReplicaCAD.

Concurrent odometry and Map-Det3D (132\,ms every ten frames) yield 26--27 FPS on two RIO10 runs (RTX 5090).

%% file: tables/replicacad_factors.tex
\begin{table*}[!t]
  \caption{Recall at 1\,m/10$^\circ$ under illumination and furniture changes in ReplicaCAD.}
  \label{tab:replicacad_factors}
  \centering
  \small
  \setlength{\tabcolsep}{4pt}
  \begin{tabular}{@{}lcccccccc@{}}
    \toprule
    & \multicolumn{4}{c}{Illumination} & \multicolumn{4}{c}{Rearrangement} \\
    \cmidrule(lr){2-5}\cmidrule(l){6-9}
    Method & day & evening & night & $\Delta_L$ & unchanged & rearranged & $\Delta_R$ & kept \\
    \midrule
    HLoc                         & 55.5 & 51.9 & 37.7 & $-17.7$ & 75.9 & 42.9 & $-33.0$ & 56\,\% \\
    ACE-G                        & 53.1 & 49.6 & 46.8 & $-6.3$  & 78.4 & 44.1 & $-34.3$ & 56\,\% \\
    R-SCoRe                      & 54.2 & 50.3 & 41.5 & $-12.6$ & 74.7 & 43.4 & $-31.3$ & 58\,\% \\
    MSG-Loc                      & 12.2 & 12.0 &  9.6 & $-2.6$  & 33.0 &  6.9 & $-26.1$ & 21\,\% \\
    \midrule
    FPFH--ICP$^\dagger$          & 10.1 &  9.1 &  8.9 & $-1.2$ & 40.1 &  3.2 & $-36.8$ &  8\,\% \\
    ACE-G$^\dagger$              & \underline{68.9} & 65.7 & \underline{66.5} & $-2.4$  & \textbf{89.4} & \underline{62.6} & $-26.8$ & 70\,\% \\
    R-SCoRe$^\dagger$            & 68.8 & \underline{66.8} & 64.6 & $-4.3$  & 87.8 & 62.5 & $-25.3$ & \underline{71\,\%} \\
    \midrule
    \method{} (ours), causal             & 60.0 & 64.2 & 59.0 & $\underline{-1.0}$  & 80.1 & 57.2 & $\underline{-22.9}$ & \underline{71\,\%} \\
    \method{} (ours), with clip closure  & \textbf{73.8} & \textbf{76.8} & \textbf{73.9} & $\mathbf{+0.1}$  & \underline{88.5} & \textbf{72.1} & $\mathbf{-16.4}$ & \textbf{81\,\%} \\
    \bottomrule
  \end{tabular}
  \par\smallskip
  \begin{minipage}{\textwidth}
  \footnotesize
  Recall (\%, higher is better) uses all frames; illumination averages six layouts per column, and unchanged/rearranged average three/fifteen tours.
  $\Delta_L$ and $\Delta_R$ are night--day and rearranged--unchanged changes (points); kept is rearranged/unchanged (\%). Rankings use unrounded recall.
  Bold/underline mark first/second; ties share rank.
  $^\dagger$ is defined as in Table~\ref{tab:main}.
  FPFH--ICP factors use per-tour medians across replay seeds; Table~\ref{tab:main} retains the original run's recall.
  \end{minipage}
\end{table*}

%% file: sections/conclusion.tex
\section{CONCLUSION}
\label{sec:conclusion}

\method{} registers adaptive, odometry-connected clips against compact class-labelled box maps for long-term indoor relocalization.
Hypothesis-first registration verifies object-triplet proposals with fused clip geometry, and orientation-aware refinement uses box faces, gravity and wall directions.
The run-level decision and clip-rigid pose graph then reconcile placements as the video arrives.
Across both datasets, this pipeline attains higher recall at 1\,m/10$^\circ$ than the evaluated per-frame scene coordinate regressors with maps roughly two orders of magnitude smaller.

\method{} assumes within-clip RGB-D odometry and class-labelled detections; its orientation fallback uses gravity and Manhattan walls when object evidence is insufficient.
Compact boxes favor robust place recovery over fine alignment after object motion.
Future work can add uncertainty-aware box updates after rearrangement while retaining compact storage.

Results on rearranged scenes show that query-side temporal evidence supports long-term relocalization with compact maps.